\documentclass{article} 
\usepackage{iclr2017_conference,times}
\usepackage{hyperref}
\usepackage{url}
\usepackage{latexsym}
\usepackage{amsmath,amsthm, amssymb, mathrsfs,empheq}
\usepackage{times}
\usepackage{algorithm}
\usepackage{algorithmic}
\usepackage{graphicx}
\usepackage{svg}
\usepackage{mathtools,bm}
\DeclarePairedDelimiterX{\inp}[2]{\langle}{\rangle}{#1, #2}
\newcommand{\beq}{\begin{equation}}
\newcommand{\eeq}{\end{equation}}
\newcommand{\beqa}{\begin{eqnarray}}
\newcommand{\eeqa}{\end{eqnarray}}
\newcommand{\ben}{\begin{enumerate}}
\newcommand{\een}{\end{enumerate}}

\newcommand{\llk}{\mathcal{L}}
\newcommand{\Real}{\mathbb{R}}

\newcommand{\lb}{\left(}
\newcommand{\rb}{\right)}
\newcommand{\ls}{\left[}
\newcommand{\rs}{\right]}

\newcommand{\vecv}{\mathbf{v}}

\newcommand{\veca}{\mathbf{a}}
\newcommand{\vecb}{\mathbf{b}}
\newcommand{\vecc}{\mathbf{c}}
\newcommand{\vecw}{\mathbf{w}}
\newcommand{\vech}{\mathbf{h}}

\newcommand{\vectheta}{\mathbf{\theta}}

\newcommand{\veclambda}{\bm{\lambda}}

\newcommand{\Ex}{\mathbb{E}}

\newcommand{\sv}{\tilde{\vecv}}

\newcommand{\st}{\tilde{\theta}}

\newcommand{\Cov}{Cov}
\newcommand{\non}{\nonumber}

\def  \wrt{\textit{w.r.t. }}

\title{Second order Learning of RBMs}
\author{Vidyadhar Upadhya}

\title{Efficient Learning of Restricted Boltzmann Machines Using Covariance Estimates}

\author{Vidyadhar Upadhya , P S Sastry\\ Indian Institute of Science Bangalore, India}

\begin{document}

\maketitle

\begin{abstract}
Learning RBMs using standard algorithms such as CD(k) involves gradient descent on the negative log-likelihood. One of the terms in the gradient, which involves expectation w.r.t. the model distribution, is intractable and is obtained through an MCMC estimate. In this work we show that the Hessian of the log-likelihood can be written in terms of covariances of hidden and visible units and hence, all elements of the Hessian can also be estimated using the same MCMC samples with small extra computational costs. Since inverting the Hessian may be computationally expensive, we propose an algorithm that uses inverse of the diagonal approximation of the Hessian, instead. This essentially results in parameter-specific adaptive learning rates for the gradient descent process
and improves the efficiency of learning RBMs compared to 
the standard methods. 
Specifically we show that using the inverse of diagonal approximation of Hessian in the stochastic DC (difference of convex functions) program approach results in 
very efficient learning of RBMs. 
\end{abstract}

\section{Introduction}
The Restricted Boltzmann Machine (RBM), an energy based generative model \citep{smolensky1986information,freund1994unsupervised,hinton2002training},
is among the basic building blocks of 
several deep learning models including Deep Boltzmann Machine (DBM) and Deep Belief Networks (DBN) \citep{salakhutdinov2009deep,Hinton06,DBLP:journals/corr/abs-1806-07066}.   
It can also be used as a discriminative model with suitable modifications. 

The traditional method of learning the parameters of an RBM involves minimizing the KL divergence between the data and the model distribution.
This is equivalent to the maximum likelihood estimation and is implemented as a gradient ascent on the log-likelihood. However, evaluating the gradient (\textit{w.r.t.} the 
parameters of the model) of the log-likelihood
is computationally expensive (exponential in minimum 
of the number of visible/hidden units in the model) since it contains an expectation 
term \textit{w.r.t.} the model distribution. Therefore, in the iterative stochastic gradient methods this term is approximated using samples from the model distribution. 
The samples are obtaining using Markov Chain Monte Carlo (MCMC) methods which are efficient in this regard due to RBM's bipartite connectivity structure.
The popular Contrastive Divergence (CD) algorithm 
uses samples obtained through an MCMC procedure with a specific initialization strategy. 
However, the resulting estimated gradient may be poor when the RBM model is high dimensional.
The poor estimate  
can make the stochastic gradient descent (SGD) based algorithms such as CD to even diverge in some cases \citep{fischer2010empirical}.

There are two general approaches to make the learning of RBMs more efficient.
The first is to design an efficient MCMC method to get good representative samples from the model distribution and 
thereby reduce the variance of the estimated
gradient \citep{desjardins2010adaptive,tieleman2009using}. However, advanced MCMC methods are computationally 
intensive, in general. The second 
approach is to design better optimization strategies which are robust to the noise in the estimated gradient~\citep{Martens2010DeepLV,metricfree,carlson2015stochastic}. 
Most approaches to design better optimization algorithms for learning RBMs are second order optimization techniques that either need approximate Hessian inverse or an estimate of the inverse Fisher matrix
(The two approaches differ for the RBM since it contains hidden units).
The Hessian-Free (H-F) algorithm \citep{Martens2010DeepLV} is an iterative procedure which approximately solves  
a linear system to obtain the curvature through matrix-vector product. In \citep{metricfree} H-F algorithm 
is used to design natural gradient descent for learning Boltzmann machines.
A sparse Gaussian graphical model is proposed in \citep{Grosse:2015} to estimate the inverse Fisher matrix
in order to devise factorized natural gradient descent procedure. All these algorithms either need additional computations 
to solve an auxiliary linear system or are computationally intensive algorithms to directly estimate the inverse Fisher matrix.

There have been attempts to exploit the fact that the RBM log-likelihood function is a difference of convex functions by
modifying the standard difference of convex programming (DCP) approach to handle the stochasticity  
\citep{pmlr-v77-upadhya17a,pmlr-v54-nitanda17a}.
The {\it stochastic- difference of convex functions programming} (S-DCP) algorithm
\citep{pmlr-v77-upadhya17a} uses only the first order derivatives of the log-likelihood and solves a series of convex optimization problems 
using constant step-size gradient descent method for a fixed number of iterations. 
The {\it Stochastic proximal DC} (SPD) algorithm \citep{pmlr-v54-nitanda17a}  
uses an additional proximal term along with the DC objective function and solves series of convex optimization problems.
Unlike S-DCP, the SPD solves each subproblem to a certain level of accuracy (predefined). In order to achieve the
required accuracy level large minibatch is used which significantly increases the computational cost \citep{pmlr-v97-xu19c}.     
However, the computational cost of S-DCP algorithm can be made identical to that of 
CD based algorithms with a proper choice of hyperparameters and is shown to perform well compared to other algorithms \citep{pmlr-v77-upadhya17a}.


Motivated by the simplicity and the efficiency of the S-DCP algorithm, in this work, we modify the S-DCP 
algorithm using the diagonal approximation of the Hessian of 
the log-likelihood and propose a diagonally scaled S-DCP, denoted as S-DCP-D. Use of a diagonal approximation of Hessian essentially amounts to having an adaptive stepsize which is different for different parameters.  

We show that the diagonal terms of the Hessian can be expressed in terms of the covariances of the visible and hidden units and can be estimated using 
the same MCMC samples that are used to get the gradient estimates. Therefore,
the additional computational cost incurred is small. Thus, the main contribution of the paper is a well-motivated algorithm (with small additional computational costs)
that can automatically adopt the step-size (through the inverse of the diagonal approximation of the Hessian) to improve the efficiency of learning an RBM.
  Through  empirical investigations we show the effectiveness of the proposed algorithm.


The rest of the paper is organized as follows. In section \ref{sec:background}, 
we briefly describe the RBM model and the maximum likelihood (ML) learning approach for RBM. 
We explain the proposed algorithm, S-DCP-D, in section \ref{sec:D-S-DCP}.
In section \ref{sec:experiments}, we present simulation results on some benchmark datasets to show the efficiency of S-DCP-D.  Finally, we conclude the paper in 
section~\ref{sec:conclusions}.

\section{Background}
\label{sec:background}
\subsection{Restricted Boltzmann Machines}
The Restricted Boltzmann Machine (RBM) is an energy based model
with a two layer architecture, in which $m$ visible stochastic units $(\vecv)$ in one layer are connected to $n$ hidden stochastic units $(\vech)$ in the other layer \citep{smolensky1986information,freund1994unsupervised,hinton2002training}. There are no 
connections from visible to visible and hidden to hidden nodes and the connections between the layers are undirected.
An RBM with parameters $\theta$ represents a probability distribution   
\beq
 p(\vecv,\vech\vert\theta)=e^{-E(\vecv,\vech;\theta)}/Z_\theta\label{llk_eq}
 \eeq
 where, $Z_\theta=\sum_{\vecv,\vech}e^{-E(\vecv,\vech;\theta)}$ is the normalizing constant which is called the 
 partition function and 
$E(\vecv,\vech;\theta)$ is the energy function.
The energy function is defined based on the type of units, discrete or continuous. 
In this work, we consider binary units, i.e., $\vecv\in\{0,1\}^m$ and $\vech\in\{0,1\}^n$ for which the 
energy function is 
defined as
 \beq
 E(\vecv,\vech;\theta)=-\sum_{i,j}w_{ij} h_i\, v_j-\sum_{j=1}^{m} b_j\,v_j-\sum_{i=1}^{n} c_i\, h_i
 \eeq
  where, $\theta=\{\vecw\in\Real^{n\times m},\vecb\in\Real^{m},\vecc\in\Real^{n}\}$ is the set of model parameters. Here,
  $w_{ij}$ is the weight of the connection between the $i^\text{th}$ hidden 
  unit and the $j^{\text{th}}$ visible unit. The $c_i$ and $b_j$ denote the bias for the $i^{\text{th}}$ hidden 
  unit and the $j^{\text{th}}$ visible unit, respectively.  
\subsection{Maximum Likelihood Learning}\label{sec_ML}
One of the methods to learn the RBM parameters, $\theta$, is through the maximization of the 
log-likelihood over the training samples. The log-likelihood, for a given training sample ($\vecv$), is given by,
\beqa
\llk (\theta\vert \vecv)&{=}&\log\,p(\vecv\vert\theta)\non
 = \log\,\sum_\vech p(\vecv,\vech\vert\theta)\non\\
& \triangleq &  g(\theta,\vecv) - f(\theta) \label{ll_base}
\eeqa
where,  
\beqa
g(\theta,\vecv)&=& \log\,\sum_\vech e^{-E(\vecv,\vech:\theta)}\non\\
f(\theta)&=& \log\,Z_\theta=\log \,\sum_{\vecv',\vech}e^{-E(\vecv',\vech;\theta)}.\label{f_g_def}
\eeqa
The optimal RBM parameters can be found by solving the following optimization problem.
\beq
\theta^*=\arg \max_\theta \llk (\theta\vert \vecv)= \arg \max_\theta \,\,( g(\theta,\vecv)- f(\theta))\label{ll_opt}
\eeq
The above optimization problem is solved using an iterative gradient ascent procedure:
\beq
\theta^{t+1}=\theta^{t}+\left.\eta\,\,\nabla_\theta\llk (\theta \vert \vecv)\right\vert_{\theta=\theta^t}\non
\eeq
The gradient of $g$ and $f$ are given by \citep{hinton2002training,fischer2012introduction},
\beqa
\nabla_\theta\,g(\theta,\vecv)&=&-\frac{\sum_\vech e^{-E(\vecv,\vech:\theta)} \nabla_\theta\, E(\vecv,\vech;\theta)}{\sum_\vech e^{-E(\vecv,\vech:\theta)}}\non\\ 
&=&-\Ex_{p(\vech\vert\vecv;\theta)}\ls \nabla_\theta\,E(\vecv,\vech;\theta)\rs\non\\
\nabla_\theta \,f(\theta)&=& -\frac{\sum_{\vecv',\vech} e^{-E(\vecv',\vech;\theta)}\nabla_\theta\, E(\vecv',\vech;\theta)}{\sum_{\vecv',\vech} e^{-E(\vecv',\vech;\theta)}} \non\\
&=&-\Ex_{p(\vecv',\vech;\theta)}\ls\nabla_\theta\, E(\vecv',\vech;\theta)\rs\label{loglik_grad}
\eeqa
where, $ \Ex_q$ denotes the expectation \wrt the distribution $q$. 
The expectation under the conditional distribution, $p(\vech\vert\vecv;\theta)$, for a given $\vecv$, has a closed form expression 
and hence, $\nabla_\theta\,g$ is easily evaluated analytically. However, expectation under 
the joint density,  $p(\vecv,\vech;\theta)$, is computationally intractable since the number of terms in the expectation 
summation grows exponentially with  
(minimum of) the number of 
hidden units/visible units present in the model. Hence, sampling methods are used to obtain $\nabla_\theta\,f$.

The 
contrastive divergence \citep{hinton2002training}, a popular algorithm to learn RBMs, uses a single sample (obtained after running a Markov chain for $K$ steps) to approximate the expectation as,
\beqa
\nabla_\theta \,f(\theta)&=&-\Ex_{p(\vecv,\vech;\theta)}\ls\nabla_\theta\, E(\vecv,\vech;\theta)\rs\non\\
&=&-\Ex_{p(\vecv;\theta)}\Ex_{p(\vech\vert\vecv;\theta)}\ls\nabla_\theta\, E(\vecv,\vech;\theta)\rs\non\\
&\approx&-\Ex_{p(\vech\vert{\sv}^{(K)};\theta)}\ls\nabla_\theta\, E({\sv}^{(K)},\vech;\theta)\rs\non\\
&\triangleq& \hat{f'}(\theta,\sv^{(K)})
\eeqa
Here, $\sv^{(K)}$ is the sample obtained after $K$ transitions of the Markov chain (defined by the current parameter values $\theta$) 
initialized with the training sample $\vecv$. 
There exist many variations of this CD algorithm in the literature, such as 
 persistent (PCD) \citep{tieleman2008training}, fast persistent (FPCD) \citep{tieleman2009using}, population (pop-CD) \citep{KrauseFI15}, average contrastive divergence (ACD) \citep{e18010035}
 and weighted contrastive divergence (WCD) \citep{DBLP:journals/corr/abs-1801-02567}. 
Another popular algorithm, parallel tempering (PT) \citep{desjardins2010adaptive}, is also based on MCMC. 
All these algorithms differ in the way they obtain representative samples from the model distribution for estimating the gradient.
The centered gradient (CG) \citep{Montavon2012} algorithm also uses the same principle as that of CD algorithm to obtain the samples;  
however, while estimating the gradient it removes the mean of the training data and the mean of the hidden activations from the 
visible and the hidden variables respectively. This approach has been seen to improve the conditioning of the underlying 
optimizing problem \citep{Montavon2012}. 

As mentioned earlier, here we propose S-DCP-D which is a modification of the S-DCP algorithm~\citep{pmlr-v77-upadhya17a}.
The S-DCP approach is advantageous since a non-convex problem is solved by 
iteratively solving a sequence of convex optimization problems. 


\section{Diagonally scaled S-DCP (S-DCP-D)}\label{sec:D-S-DCP}
The DCP \citep{yuille2002concave,An2005} is an algorithm useful for solving optimization problems of the form,
\beq
\theta^{*}=\arg \min_\theta F(\theta)=\arg \min\limits_\theta \lb f(\theta)-g(\theta)\rb\label{CCP_DCA_base}
\eeq
where, both the functions $f$ and $g$ are convex and smooth but $F$ is non-convex. 
It is an iterative procedure defined by,
\beq
\theta^{(t+1)}=\arg \min_\theta \lb f(\theta)-\theta^T \nabla g(\theta^{(t)})\rb.\label{CCP_DCA_convx}
\eeq
In the RBM setting, $F$ corresponds to the negative log-likelihood function and the functions $f,g$ are as defined in eq.~(\ref{f_g_def}). 

In the S-DCP algorithm, the convex optimization problem given by RHS of  
eq.~(\ref{CCP_DCA_convx}) is (approximately) solved  using a few iterations of gradient descent on $f(\theta)-\theta^T \nabla g(\theta^{(t)},\vecv)$  for which
the $\nabla f$ is estimated using samples obtained though MCMC (as in Contrastive Divergence). Thus, it is a stochastic gradient descent for the (convex) objective 
function $f(\theta)-\theta^T \nabla g(\theta^{(t)},\vecv)$ for a fixed number of iterations (denoted as $d$). 
A description of this S-DCP algorithm is given as  
Algorithm \ref{S-DCP}. Note that, it is possible to choose the hyperparameters $d$ and $K'$ such that the amount of 
computation required is identical to CD($K$) algorithm~\citep{pmlr-v77-upadhya17a}.

\begin{algorithm}[bt]
   \caption{S-DCP update for a single training sample $\vecv$}\label{S-DCP}
\begin{algorithmic}
   \STATE {\bfseries Input:} $\vecv,\theta^{(t)},\eta,d, K'$
   \STATE Initialize $\st^{(0)}=\theta^{(t)},\sv^{(0)}=\vecv$
       \FOR{$l=0$ {\bfseries to}  $d-1$}
           \FOR{$k=0$ {\bfseries to}  $K'-1$}
               \STATE sample $h_i^{(k)}\sim p(h_i\vert\sv^{(k)},\st^{(l)}), \forall i$
	       \STATE sample $\tilde{v}_j^{(k+1)}\sim p(v_j\vert\vech^{(k)},\st^{(l)}), \forall j$
           \ENDFOR
           \STATE $\st^{(l+1)}=\st^{(l)}-\eta \ls\hat{f'}(\st^{(l)},\sv^{(K')})-\nabla g (\theta^{(t)},\vecv)\rs $  
           \STATE $\sv^{(0)}=\sv^{(K')}$\label{v0vk}
   \ENDFOR
\STATE {\bfseries Output:}  $\theta^{(t+1)}=\st^{(d)}$
   \end{algorithmic}
\end{algorithm}

The S-DCP algorithm can be viewed as two loops. The outer loop is the iteration given by eq.~(\ref{CCP_DCA_convx}). Each iteration here involves a convex optimization which is (approximately) solved by the inner loop of S-DCP through stochastic gradient descent (w.r.t. $\theta$) on the convex function, $f(\theta)-\theta^T \nabla g(\theta^{(t)})$. 
 The proposed S-DCP-D is a scaling of this stochastic gradient descent by using the diagonal elements of the Hessian of this convex function. 

The Hessian of the objective function $f(\theta)-\theta^T \nabla g(\theta^{(t)})$ can be obtained as,
\begin{align}
& \nabla_\theta^2 f(\theta) =  - \nabla_\theta \Ex_{p(\vecv,\vech;\theta)}\ls\nabla_\theta\, E(\vecv,\vech;\theta)\rs\non\\
=&- \sum\limits_{\vecv,\vech} \nabla_\theta E(\vecv,\vech) \,\,\nabla_\theta p(\vecv,\vech;\theta)^T\,\,(\text{Since } \nabla_\theta^2 E(\vecv,\vech){=}0)\non\\
=&- \sum\limits_{\vecv,\vech} \frac{\nabla_\theta E(\vecv,\vech)}{Z_\theta^2} e^{-E(\vecv,\vech)} ( -Z_\theta\, \nabla_\theta E(\vecv,\vech)^T - \nabla_\theta Z_\theta^T)\non\\
=&- \sum\limits_{\vecv,\vech}  \nabla_\theta E(\vecv,\vech) \ls - \nabla_\theta E(\vecv,\vech)^T - \nabla_\theta \log Z_\theta^T \rs p(\vecv,\vech)\non
\end{align}
By substituting $\nabla_\theta \log Z_\theta=-\Ex_{p(\vecv',\vech';\theta)}\ls\nabla_\theta\, E(\vecv',\vech';\theta)\rs$ from eq. \eqref{loglik_grad} in the above equation, we get,
\beq
\nabla_\theta^2 f(\theta)= \Cov_{p(\vecv,\vech)}\ls \nabla_\theta E(\vecv,\vech),\nabla_\theta E(\vecv,\vech) \rs\label{Cov_n}
\eeq
where, $\Cov_q(X, X) = \Ex_q(XX^T) - \Ex_q(X) \Ex_q(X^T)$. 

Note that a typical element in $\nabla_\theta^2 f$ is $\frac{\partial^2 f}{\partial \theta_i \partial \theta_j}$ where $\theta_i$ refers to the parameters of the RBM, namely, all the $w_{ij}, b_i, c_j$. 
The diagonal element corresponding to $w_{ij}$ is
\beqa
\frac{\partial^2 f}{\partial w_{ij} \partial w_{ij}} &{=}& \Ex_{p(\vecv,\vech)} \ls\lb \frac{\partial f}{\partial w_{ij}}\rb^2 \rs - \lb \Ex_{p(\vecv,\vech)}\ls \frac{\partial f}{\partial w_{ij}} \rs\rb^2\non\\
& =& \Ex_{p(\vecv,\vech)}\ls v_j^2 \, h_i^2\rs  - \lb\Ex_{p(\vecv,\vech)}\ls v_j \, h_i\rs\rb^2\non\\
 &=&\Ex_{p(\vecv,\vech)}\ls v_j \, h_i\rs-\lb\Ex_{p(\vecv,\vech)}\ls v_j \, h_i\rs\rb^2\non\\
 &=&\Ex_{p(\vecv,\vech)}\ls v_j \, h_i\rs \lb 1-\Ex_{p(\vecv,\vech)}\ls v_j \, h_i\rs\rb\non
  \eeqa
We have used the property that $v_j^2=v_j$ and  $h_i^2=h_i$ (since $v_j,h_i$  are binary random variables) in the above derivation. 
Similarly, the diagonal terms corresponding to the bias terms are given by,
\beqa
\frac{\partial^2 f}{\partial b_{j} \partial b_{j}} & =& \Ex_{p(\vecv,\vech)}\ls  v_j\rs-\lb\Ex_{p(\vecv,\vech)}\ls \, v_j\rs\rb^2\non\\
\frac{\partial^2 f}{\partial c_{i} \partial c_{i}} & =& \Ex_{p(\vecv,\vech)}\ls  h_i\rs-\lb\Ex_{p(\vecv,\vech)}\ls \, h_i\rs\rb^2\non
\eeqa
By using the above equations, the diagonal elements of the Hessian of $f$ can be estimated simply by using the same MCMC samples used for 
gradient estimates.
For a compact notation, the diagonal terms in $\nabla_\theta^2 f$  can be written as
\beq
\text{Diag}\lb\nabla_\theta^2 f\rb =  -\Ex_{p(\vecv,\vech)} \ls\nabla_\theta E(\vecv,\vech)\rs \bm{\odot} \lb \mathbf{1}+\Ex_{p(\vecv,\vech)}\ls\nabla_\theta E(\vecv,\vech)\rs\rb\non
\eeq
where $\bm{\odot}$ represents element-wise multiplication, $\mathbf{1}$ represents vector of all ones and Diag$(\bm{A})$ represents the vector consisting of the diagonal elements of matrix $\bm{A}$.
These estimates are used in obtaining the gradient descent updates (in the inner loop S-DCP) as,
\beqa
\vectheta_l^{t+1}= \theta_l^{t} -\eta \left. \frac{\ls\nabla_\theta \lb f(\theta)-\theta^T \nabla g(\theta^{(t)})\rb\rs_l}{\ls H_t +\epsilon\mathbf{I})\rs_l}\right\rvert_{\theta=\theta^t}
\eeqa
where $\ls\veca\rs_l$ represents $l^{\text{th}}$ element of vector $\veca$, $\mathbf{I}$ is the identity matrix of appropriate dimension, $H_t$ is the estimated Hessian at iteration $t$ and $\epsilon$ is a small constant (and the term $\epsilon I$ is added for numerical stability). A detailed description of the proposed algorithm is given as  
Algorithm \ref{SDCP_minibatch}.

The inverse of the diagonal approximation of the Hessian essentially provides parameter-specific learning rates for the gradient ascent process.
In case of S-DCP algorithm the objective function for the gradient descent is convex and  the diagonal terms of the $\nabla^2 f$ are greater than or equal to zero since $f$ is convex. 
Therefore, inverse of the diagonal terms of the Hessian added with a small $\epsilon$ is numerically stable. 

Since the estimate of the gradient is noisy, the estimated Hessian is also noisy. Therefore, exponential averaging of the estimated Hessian is
used to make the algorithm stable in terms of learning. Let $\tilde{H}_t$ denote the $\nabla_\theta^2 f$ calculated at iteration $t$ as explained earlier. Let $H_t$ denote the Hessian that is used at iteration $t$ for updating the weights. We calculate $H_t$ as 
\beq
H_t= \lambda_H H_{t-1} + (1-\lambda_H) \tilde{H}_t
\eeq
where $\lambda_H$ is a parameter that decides the memory of the exponential averaging.

\begin{algorithm}[tb]
   \caption{S-DCP-D update for a mini-batch of size $N_B$}\label{SDCP_minibatch}
\begin{algorithmic}
   \STATE {\bfseries Input:} $V=[\vecv^{(0)},\vecv^{(1)},\ldots,\vecv^{(N_B-1)}],\theta^{(t)},\eta,d,K',\epsilon$
   \STATE Initialize $\st^{(0)}=\theta^{(t)},V_T=V$
       \FOR{$l=0$ {\bfseries to}  $d-1$}
       \STATE $\Delta\theta=\boldsymbol{0}, G_f=\boldsymbol{0}$
        \FOR{$i=0$ { \bfseries to} $N_B-1$}
\STATE $\sv^{(0)}=V_T[:,i]\quad\rightarrow[i^{\text{th}}\text{ column of } V_T]$
           \FOR{$k=0$ {\bfseries to}  $K'-1$}
               \STATE sample $h_i^{(k)}\sim p(h_i\vert\sv^{(k)},\st^{(l)}), \forall i$
	       \STATE sample $\tilde{v}_j^{(k+1)}\sim p(v_j\vert\vech^{(k)},\st^{(l)}), \forall j$
           \ENDFOR
           \STATE $ \Delta \theta=\Delta \theta+ \ls\hat{f'}(\st^{(l)},\sv^{(K')})-\nabla g (\theta^{(t)},\vecv^{(i)})\rs$
           \STATE $G_f \,=\, G_f\,+ \hat{f'}(\st^{(l)},\sv^{(K')})$
           \STATE $V_T[:,i]=\sv^{(K')}$
            \ENDFOR
            \STATE $H_f = \frac{G_f}{N_B} \odot \lb 1-\frac{G_f}{N_B}\rb\quad $ /* $\odot\quad$ represents element-wise multiplication     */
           \STATE $\st_s^{(l+1)}=\st_s^{(l)}-\frac{\eta}{H_{f_s}+\epsilon} \frac{\Delta \theta_s}{N_B}, \,\, \forall s$  /* $H_{f_s}$ is the diagonal element corresponding to $\theta_s$ */
   \ENDFOR
\STATE {\bfseries Output:}  $\theta^{(t+1)}=\st^{(d)}$
   \end{algorithmic}
\end{algorithm}

\subsection{Computational Complexity}\label{subsec_CC}
The computational cost of the CD$(K)$ algorithm for a mini-batch of size $N_B$ is $(N_B(KT+2L))$
where $T$ is the cost for one Gibbs transition and $L$ is the cost for evaluating $\nabla g$ (and also $\hat{f'}$).
The S-DCP algorithm with $K'$ MCMC transitions and $d$ inner loop iterations has cost
$(d\, N_B(K' T+L)+N_B L)$. The computational cost of CD($K$) and S-DCP are identical if $K'$ and $d$ are chosen to satisfy $KT=dK'T+(d-1)L$\citep{pmlr-v77-upadhya17a}. (By choosing $K = dK'$ we can make the computational costs of the two algorithms nearly equal). The difference between S-DCP and S-DCP-D is only in terms of estimating the diagonal terms of the Hessian.  
An additional $d (mn+m+n)$ elementwise multiplications are required to obtain the estimate of the the diagonal of Hessian. This represents the additional computational cost of S-DCP-D compared to that of S-DCP. 
\section{Experiments and Discussions}\label{sec:experiments}
In this section, we give a detailed comparison between the S-DCP-D and other algorithms, namely, 
centered gradient (CG)  \citep{JMLR:v17:14-237}, S-DCP, CD and PCD algorithms.
The CG algorithm is essentially a CD($k$) algorithm with additional centering heuristic which improves learning. 
Further, the objective here is to compare algorithms which have similar computational complexity and hence we do not
consider algorithms which are significantly computationally expensive (SPD, H-F, etc).

\subsection{The Experimental Set-up}\label{subsec:exp_setup}
We consider four benchmark datasets in our analysis, namely, Bars \& Stripes \citep{mackay2003information}, MNIST\footnote{statistically binarized as in \citep{salakhutdinov2008quantitative}} \citep{LeCun:1998}, CalTech $101$ Silhouettes DataSet \citep{caltecch} and \textit{kannada}-MNIST \citep{prabhu2019kannada}.
The Bars \& Stripes dataset of data dimension $D\times D$ is generated using a two-step procedure.
In the first step, all the pixels in each row are set to zero or one with equal probability and then the 
pattern is rotated by $90$ degrees with a probability of $0.5$ in the second step. We have choose  $D=3$, for which we get
$16$  distinct patterns. 
We refer to
MNIST, CalTech and the \textit{kannada}-MNIST datasets as large datasets. The MNIST , CalTech $101$ Silhouettes and the \textit{kannada}-MNIST datasets have data dimension of $784$.  

For the Bars \& Stripes dataset, we consider three RBMs with $4,8,16$ hidden units and for the large datasets, we consider RBMs with $500$
hidden units.   
We evaluate the algorithms using the performance measures obtained from multiple trials, where each trial 
fixes the initial configuration of the weights and biases. The biases of visible units and hidden units are initialized to
the inverse sigmoid of the training sample mean and zero, respectively. 
The weights are initialized to samples drawn from a Gaussian distribution 
with mean zero and standard deviation $0.01$. 
We use $25$ trials for the Bars \& Stripes dataset and $10$ trials for the large datasets.
The mini-batch learning procedure is used and the training dataset is shuffled after every epoch. 
However, for Bars $\&$ Stripes dataset full batch
training procedure is used.
We learn the RBM for a fixed number of epochs and avoid using any stopping criterion. The training is 
performed for $5000$ epochs for Bars \& Stripes dataset (corresponding to $5000$ gradient updates, due to full batch training) and $200$ epochs for the MNIST dataset (corresponding to $60,000$ gradient updates
due to the batch size of $200$).

We compare the performance of the proposed S-DCP-D with centered gradient (CG),S-DCP, CD and PCD.
We keep the computational complexity (on each mini-batch) of S-DCP roughly the same as that of CD by  
choosing $K,d$ and $K'$ such that $K=dK'$~\citep{pmlr-v77-upadhya17a}. 
Since previous works
stressed on the necessity of using a large $K$ for CD based algorithms to get a sensible generative model \citep{carlson2015stochastic,salakhutdinov2008quantitative}, 
we use $K=24$ in CD (with $d=6,K'=4$ for S-DCP) for large datasets and $K=4$  in CD (with $d=2,K'=2$ for S-DCP) for Bars \& Stripes dataset. 
In order to get an unbiased comparison, we did not use momentum and weight decay for any of the algorithms.
For the centered gradient algorithm, we use the Algorithm $1$ in \citep{JMLR:v17:14-237} which corresponds to $dd_s^b$ in their notation.
We use CD step size $K=24$ and the hyperparameters $\nu_\mu$ and $\nu_\lambda$
are set to $0.01$. The initial value of $\mu$ is 
set to the mean of the training data and $\veclambda$ is set to $\mathbf{0.5}$.

The learning rate and other hyperparameters for each algorithm is set to obtain the best performance by doing a grid search over a set of values of hyperparameters. 

\subsection{Evaluation Criterion}\label{subsec:Eval}
The performance comparison is based on the log-likelihood achieved on the training and test samples. 
For comparing the speed of learning of different algorithms, the average train log-likelihood is a reasonable measure. The average test log likelihood also indicates how well the learnt model generalizes. 
We show the maximum (over all trials) of the average train and test 
log-likelihood. 
The average test log-likelihood (denoted as ATLL) is evaluated as,
\beq
ATLL=\frac{1}{N}\sum_{i=1}^N \log \, p(\vecv_\text{test}^{(i)}\vert\theta)
\eeq
We evaluate the average train log-likelihood similarly by using the training samples rather than the test samples.
For small RBMs the above expression can be evaluated exactly.
However for large RBMs, we estimate the ATLL with annealed importance sampling \citep{neal2001annealed}
with $100$ particles and $10000$ intermediate distributions according to a linear temperature scale between $0$ and $1$.

The evaluation in terms of the generative ability of the learnt models is 
carried out by observing the samples that they generate.
We randomly initialize the states of the visible units and run the alternating Gibbs Sampler for $5000$ steps (for large datasets)/$200$ steps (for Bars \& Stripes dataset) and plot the state of the visible units. 

Overall, we use three evaluation criteria to show the effectiveness of the proposed S-DCP-D algorithm, specifically, i) speed of convergence
ii) generalization (Average Test log-likelihood) and iii) generative ability (quality of the generated samples).
\subsection{Performance Comparison}
In this section, we present  experimental results to illustrate
the performance of S-DCP-D in comparison with the other algorithms (CG, S-DCP, CD and PCD).
The algorithms are implemented using Python and CUDAMat (A CUDA-based matrix class for Python bindings)\citep{Mnih2009CUDAMatAC} on a system with Intel processor $i7-7700$ ($4$ CPU cores and processor base frequency $3.60$ GHz), NVIDIA Titan X Pascal GPU
and $16$ GB RAM configuration.

In our results we show that the speed of learning, in terms of number of training epochs,  exhibited by S-DCP-D is significantly higher compared to the other algorithms. As mentioned earlier, all three algorithms have comparable computational load (per minibatch) and hence comparison in terms of number of epochs would be similar to comparison in terms of actual running time. However, since the computations performed by the different algorithms are not identical, we need to understand difference in computational time per epoch of different algorithms as well. For this, we present below the actual computational time of different algorithms for a fixed number of epochs. 

The mean and standard deviation($\sigma$) of the utilized system time in seconds, for $5000$ epochs of learning for Bars \& Stripes dataset and for $200$ epochs of learning for large datasets, for each algorithm 
over $10$ trials are shown in the table below.
 \begin{table}[ht!]                                                                                          
\centering                             
\caption{The system time statistics for Bars \& Stripes and large datasets. The mean and standard deviation of system time (in seconds) is evaluated over $10$ trials.}\label{tab:time_compare}
\begin{tabular}{|c|cc|cc|}
\hline
    Algorithm & \multicolumn{2}{|r}{Bars \& Stripes} & \multicolumn{2}{|r|}{MNIST/CalTech/\textit{kannada}-MNIST}\\
    \hline
    & Mean & $\sigma$
    & Mean & $\sigma$   \\
    CD/PCD &  $1.41$ & $0.02$ & $248.15$ & $0.16$\\
 \hline
     CG &  $1.79$ & $0.07$ & $309.51$ & $0.17$\\
 \hline
 S-DCP &  $1.73$ & $0.06$  & $317.87$ & $0.39$ \\
 \hline
 S-DCP-D &  $1.92$ & $0.02$  & $385.65$ & $0.17$\\
 \hline
\end{tabular}
\end{table}

As can be seen from table~\ref{tab:time_compare}, the computational time for S-DCP is $3\%$ (for large datasets) more compared to that of CG. As mentioned earlier, by taking $K = dK'$ we can make the computational time of these two algorithms nearly same. Compared to S-DCP, the time for S-DCP-D is about $8\%$ more for Bars \& Stripes and $24\%$ more for MNIST/CalTech. The additional computation for S-DCP-D is calculating diagonal of Hessian and this grows linearly with $m, n$. 

In all results presented here we show evolution of ATLL with number of epochs for different algorithms.\footnote{Since we do not employ any stopping criterion, we cannot give `time taken to learn' for different algorithms; we can only show how log likelihood evolves with number of training epochs.} As would be seen from the results, the S-DCP-D is faster in terms of number of epochs by much more than 25\% thus justifying the claim that it results in efficient learning. In addition, on large datasets, the ATLL achieved by S-DCP-D is also larger.  


\subsubsection{Bars \& Stripes}
\begin{figure}
\centering
 \includegraphics[width=0.85\textwidth]{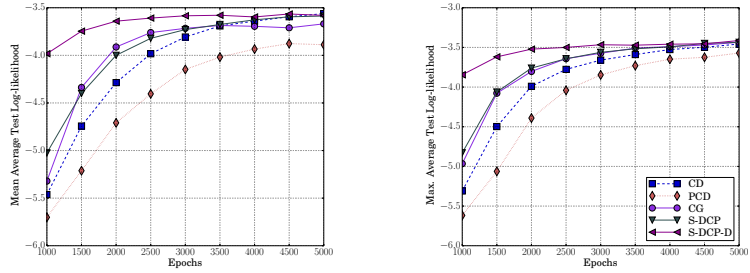}
 \caption{The evolution of mean and maximum average log-likelihood acheived on the training and the test set over all the trials on the Bars \& Stripes dataset. 
}\label{fig:small_tll}
\end{figure}

Fig. \ref{fig:small_tll} shows the evolution of the mean and maximum ATLL achieved by the 
RBM with $4$ hidden units, learnt for the Bars \& Stripes dataset. (Note that here all patterns are used for training and hence there is no distinction between training and test data sets).
As can be seen, the S-DCP-D has significantly higher speed of learning compared to S-DCP indicating the effectiveness of the parameter-specific learning rate induced by the diagonal scaling. It is also faster than CG, CD and PCD. This increased speed does not come at the expense of accuracy; the final ATLL of all algorithms is roughly same though S-DCP and CG take more epochs to converge. 
We observed similar behavior with RBM models having number of hidden units $8$ and $16$.

%


\subsubsection{Large Datasets}
\begin{figure}
\centering
 \includegraphics[width=0.99\textwidth]{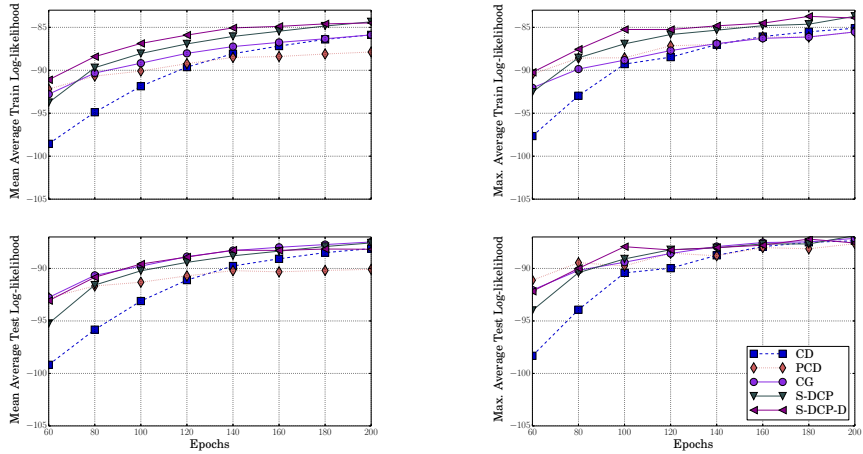}
\caption{The mean and maximum average log-likelihood over all the trials on the 
training and test set for MNIST  dataset.
Note that
the learning rate for each algorithm is set to obtain the best performance.}\label{fig:mnist_atll}
\end{figure}

\begin{figure}
\centering
 \includegraphics[width=0.99\textwidth]{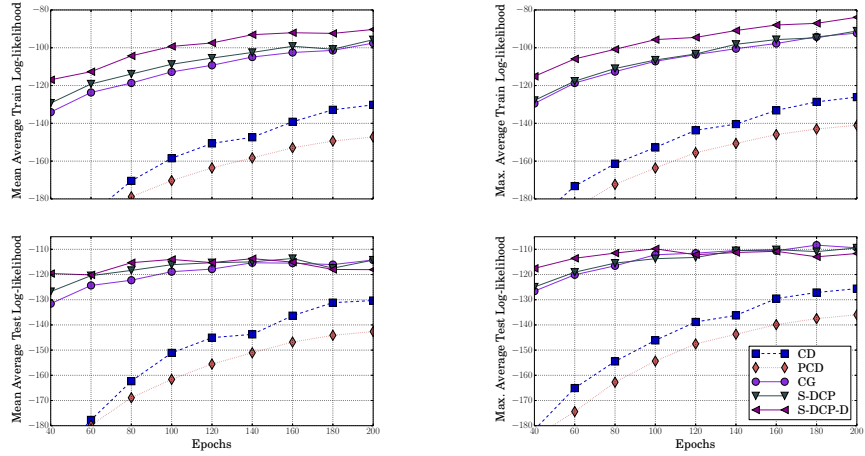}
\caption{The mean and maximum average log-likelihood over all the trials on the 
training and test set for CalTech dataset.
}\label{fig:caltech_atll}
\end{figure}

\begin{figure}
\centering
 \includegraphics[width=0.99\textwidth]{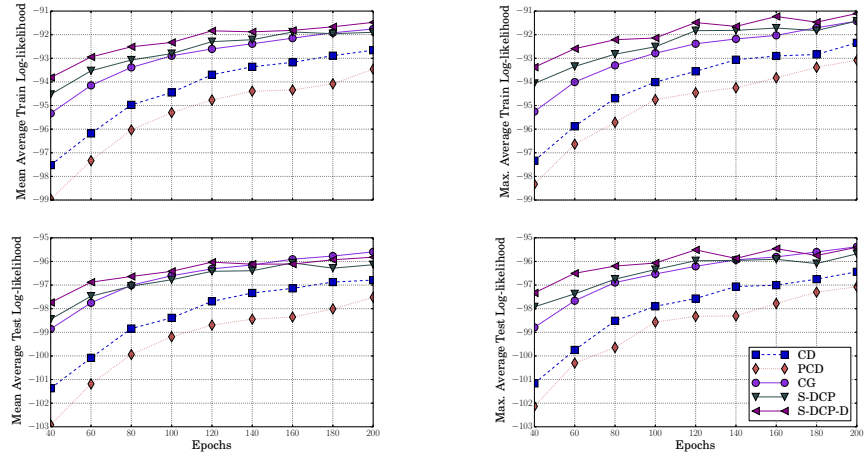}
\caption{The mean and maximum average log-likelihood over all the trials on the 
training and test set for \textit{kannada}-MNIST dataset.
}\label{fig:kmnist_atll}
\end{figure}

Fig. \ref{fig:mnist_atll}, \ref{fig:caltech_atll},\ref{fig:kmnist_atll} show the evolution of the mean and maximum average log-likelihood of the test and training set for the MNIST, CalTech and \textit{kannada}-MNIST datasets respectively.
The convergence of S-DCP-D  is faster compared to both S-DCP and CG.
We observe in Fig.  \ref{fig:caltech_atll} that the S-DCP-D evolution is smoother compared to S-DCP which
suggests that the stability of the learning algorithm is improved by the parameter-specific learning rate employed.
Further, the ATLL evolution in Fig.  \ref{fig:mnist_atll} indicates that the generalization ability of the model learnt using S-DCP-D is comparable to that learnt by the other algorithms.
The maximum ATLL achieved by S-DCP-D is $-87.1$ which is comparable to the other algorithms. 
The provided maximum ATLL score for S-DCP matches with that reported in an earlier study in \citep{pmlr-v77-upadhya17a}.
Also,
the ATLL achieved by the learnt models are comparable to that of the VAE (Variational Autoencoder) and IWAE (Importance Weighted Autoencoder) models \citep{DBLP:journals/corr/BurdaGS15}.
We observe a similar behaviour for the CalTech and \textit{kannada}-MNIST datasets, as shown in Fig. \ref{fig:caltech_atll} and \ref{fig:kmnist_atll} respectively. The performance of S-DCP-D is superior to that of S-DCP, CG, CD and PCD algorithms. 


\begin{figure}%
\subfloat[CG]{%
\includegraphics[width=0.3\columnwidth]{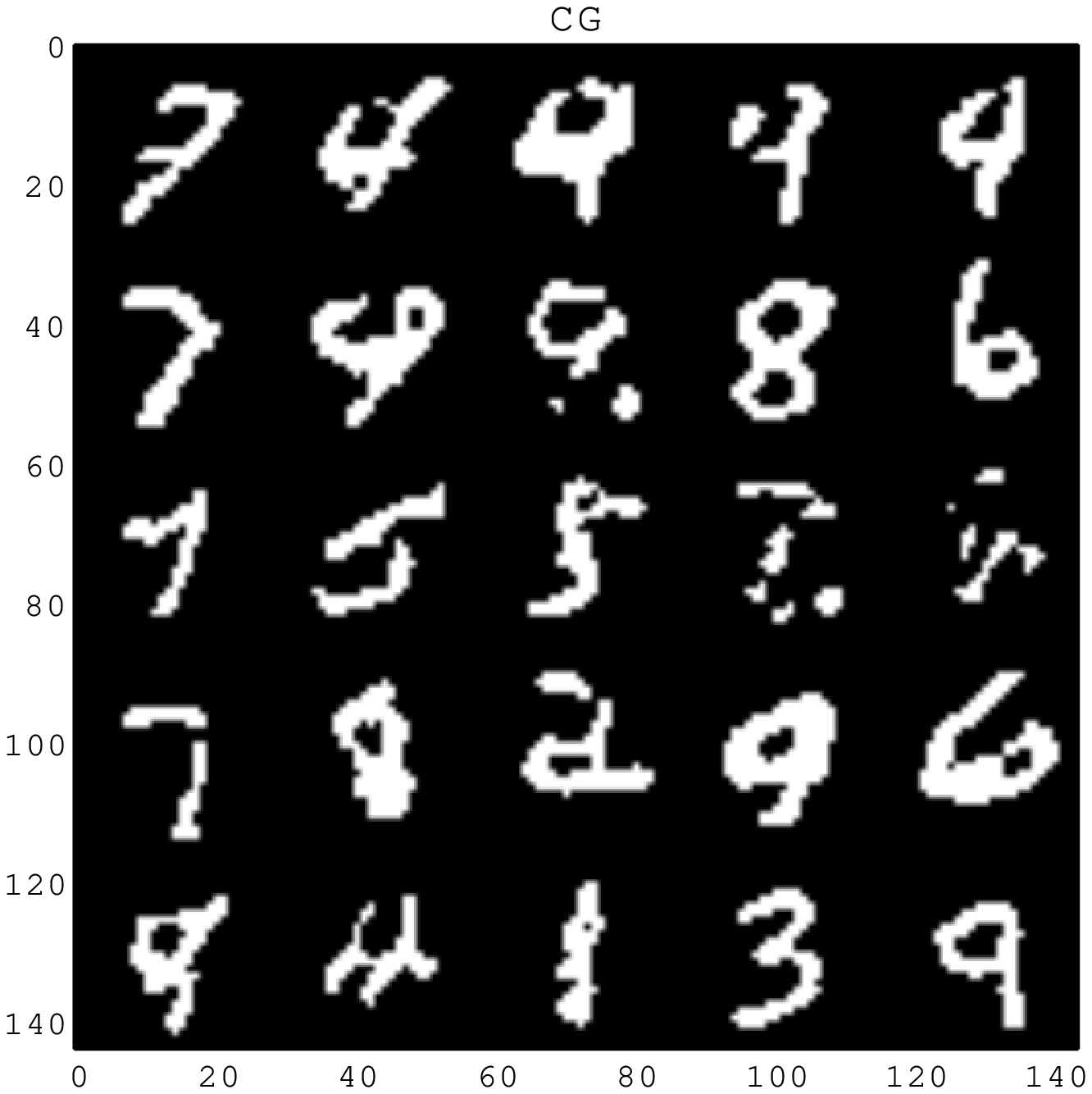}
}%
\subfloat[S-DCP]{%
\includegraphics[width=0.3\columnwidth]{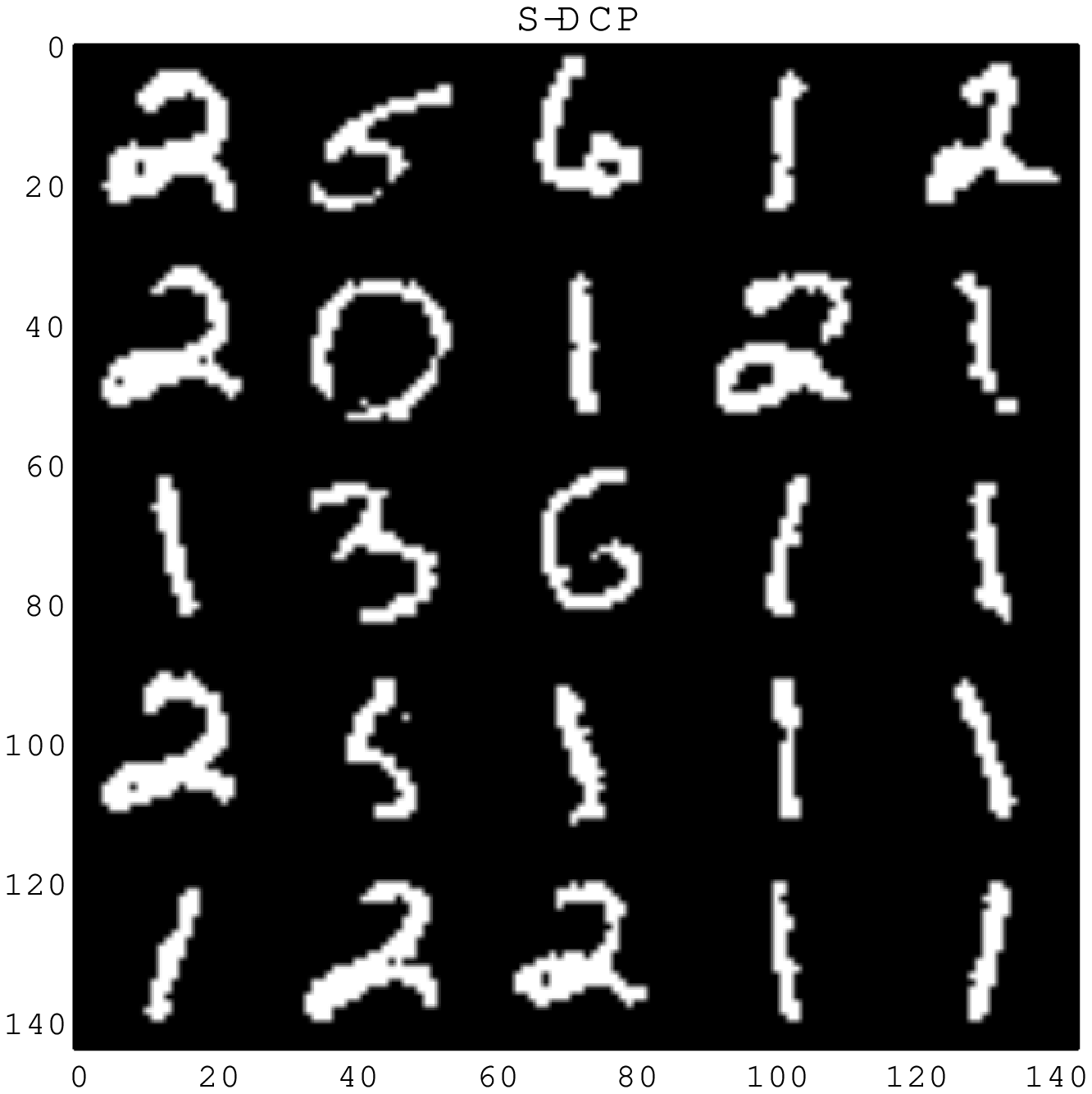}
}%
\subfloat[S-DCP-D]{%
\includegraphics[width=0.33\columnwidth]{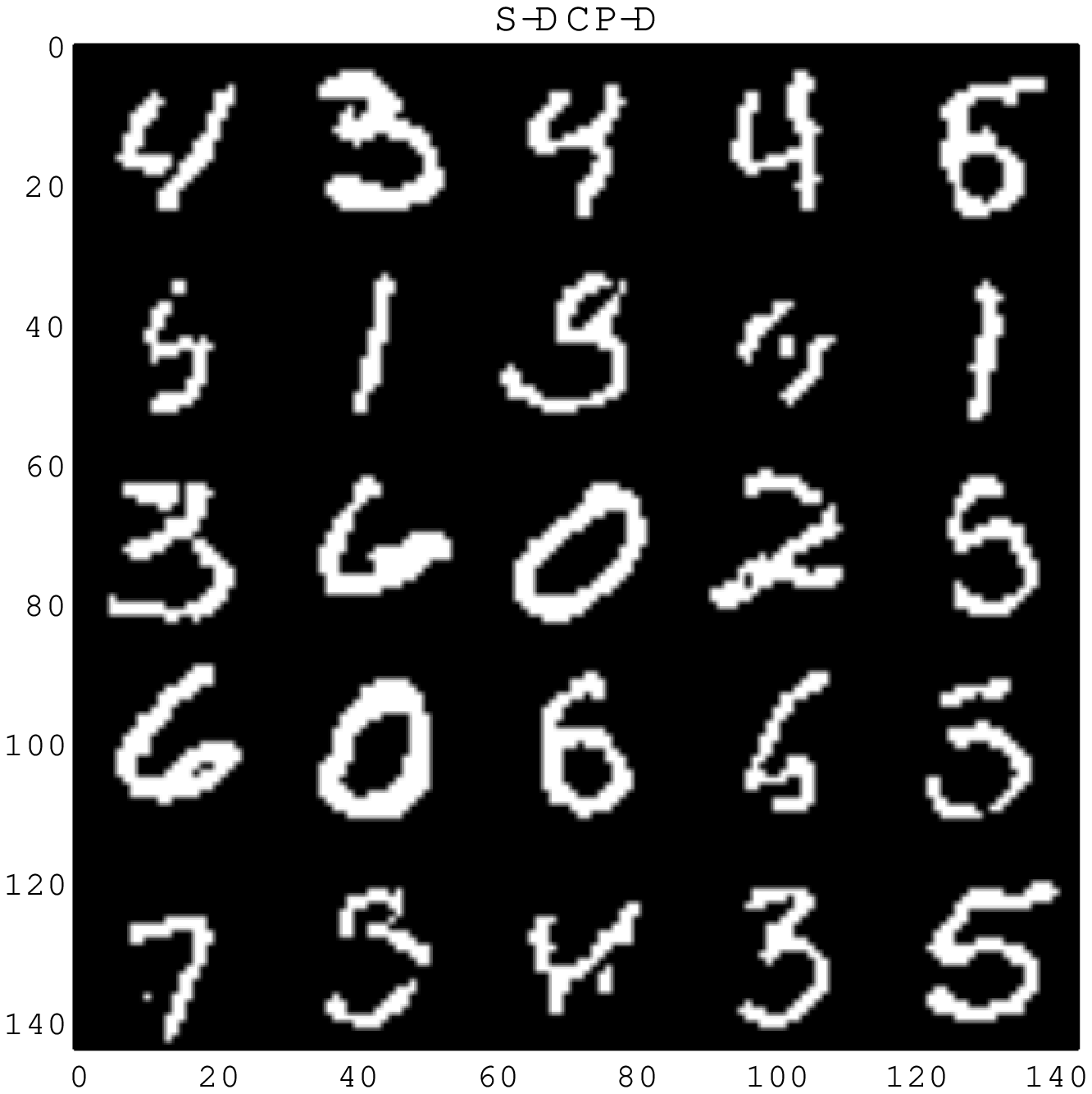}
}%

\caption{$25$ sample images generated from the models learnt on the MNIST dataset. The visible states are randomly initialized and the Gibbs sampler is run
for $5000$ steps. The final states of the visible units are shown.}\label{mnist_samples_gen}
\end{figure}

%
The samples generated by the models learnt using MNIST dataset are given in Fig.~\ref{mnist_samples_gen}. As observed from Fig.~\ref{mnist_samples_gen}, the samples generated by S-DCP-D 
are sharp compared to those produced by CG based model. Also, it can be observed that the samples generated by CG and S-DCP-D are more diverse
compared to those produced by S-DCP. We observed a similar behaviour for the CalTech and \textit{kannada}-MNIST dataset. While subjectively the samples produced by S-DCP-D look better, it is important to note that there exist no objective measures to evaluate a generative model based on 
the quality of the generated samples.
\section{Conclusions}\label{sec:conclusions}

Learning an RBM is difficult due to the noisy estimates of the gradient of the log-likelihood obtained through an MCMC procedure. In this work we proposed an algorithm where we can automatically obtain different adaptive step-sizes for gradient descent for different parameters. This is done by using the inverse of the diagonal approximation of the Hessian. We showed that the Hessian of the log likelihood is given by covariances of the model distribution and hence the Hessian can be estimated using the same MCMC samples that are used for estimating the gradient. Thus, estimating the diagonal of the Hessian has only small additional computational cost. 

Through extensive simulations, we showed that the S-DCP-D 
results in a more efficient learning of RBMs compared to S-DCP and CG algorithms. The additional attraction in using the Hessian here is that in S-DCP-D the gradient descent in the inner loop is on a convex objective function. The diagonal scaling also seems to stabilize the learning and the resulting generative model seems to produce better samples as we showed empirically.

It is known that learning of RBMs can be more efficient if the learning rate is reduced with iterations using a heuristically devised schedule. But the schedule has to be fixed through cross validation.  
  The proposed approach automatically provides 
parameter-specific learning rates which makes the learning procedure both stable and efficient. The only hyper parameters of the proposed algorithm is $\epsilon$ which does not affect the learning dynamics much and is there only to control numerical underflows.  
The main attraction of S-DCP-D, in our opinion, is its simplicity compared to other sophisticated second-order optimization techniques 
which use computationally intensive algorithms to estimate the inverse of the Hessian. 

For learning an RBM, the centered gradient algorithms are shown to be better compared to CD(k) type algorithm. The reason is conjectured to be the similarity among the second order optimization algorithms and centered gradient method. We feel that the well-motivated and simple second-order algorithm proposed, namely S-DCP-D, can provide the correct platform to further explore this view of centered gradient algorithms.

\bibliographystyle{iclr2017_conference}
\bibliography{rbm_2nd_order_arxiv.bib}
\end{document}